\documentclass[conference]{IEEEtran}
\IEEEoverridecommandlockouts
\usepackage{cite}
\usepackage{amsmath,amssymb,amsfonts}
\usepackage{algorithmic}
\usepackage{graphicx}
\usepackage{textcomp}
\usepackage{xcolor}
\usepackage{url} 
\def\BibTeX{{\rm B\kern-.05em{\sc i\kern-.025em b}\kern-.08em
    T\kern-.1667em\lower.7ex\hbox{E}\kern-.125emX}}

\def\ie{\emph{i.e.}}

\begin{document}

\title{DeepFake-o-meter: An Open Platform for DeepFake Detection}
\author{Yuezun Li$^{1\dagger}$\thanks{$\dagger$ The work was done when the author was a post-doc at University at Buffalo.}, Cong Zhang$^2$, Pu Sun$^2$, Honggang Qi$^2$, and Siwei Lyu$^3$ \\
$^1$Ocean University of China, China \\
$^2$University of Chinese Academy of Sciences, China \\
$^3$University at Buffalo, State University of New York, USA}

\maketitle

\begin{abstract}
In recent years, the advent of deep learning-based techniques and the significant reduction in the cost of computation resulted in the feasibility of creating realistic videos of human faces, commonly known as DeepFakes. The availability of open-source tools to create DeepFakes poses as a threat to the trustworthiness of the online media. In this work, we develop an open-source online platform, known as DeepFake-o-meter, that integrates state-of-the-art DeepFake detection methods and provide a convenient interface for the users. We describe the design and function of DeepFake-o-meter in this work.
\end{abstract}


\section{Introduction}

The buzzword {\tt DeepFakes} has been frequently featured in the news and social media to refer to realistic impersonating images, videos, and audios that are generated using AI algorithms. Although fabrication and manipulation of digital media are not a new phenomenon \cite{farid08}, powerful AI technology, in particular, deep neural networks (DNNs), and the unprecedented computing power have made it easier than ever to create sophisticated and compelling fakes.Left unchecked, DeepFakes can escalate the scale and danger of disinformation, and fundamentally erode our trust in digital media.

The mounting concerns over the negative impacts of DeepFakes have spawned an increasing interest in DeepFake detection. In less than three years, there have been numerous new detection methods of DeepFakes. However, differences in training datasets, hardware, and learning architectures across research publications make rigorous comparisons of different detection algorithms challenging. At the same time, the cumbersome process of downloading, configuring, and installing of individual detection algorithms deny the access of the state-of-the-art DeepFake detection methods to most users. To this end, we have developed an online DeepFake detection platform. It serves three purposes. 
\begin{itemize}
    \item For developers of DeepFake detection algorithms, it provides an API architecture to wrap individual algorithms and run on a third-party remote server.
    \item For researchers, it is an evaluation/benchmarking platform to compare multiple algorithms on the same input. 
    \item For users, it provides a convenient portal to use multiple state-of-the-art detection algorithms.  
\end{itemize}
Currently we have incorporated 10+ state-of-the-art DeepFake image and video detection methods, and will keep adding more capacities.

In this work, we describe the design and the underlying mechanism of DeepFake-o-meter in details. We start with the overall architecture of the system, which is composed of a web-based front-end that interacts with the user, and an on-server back-end to perform analyses on the input videos. The separation of front-end and back-end is to ensure the security of the user-uploaded data, as well as to accommodate the long running time and the short response time to the users. We further provide an overview of the DeepFake detection algorithms that have been integrated into the current DeepFake-o-meter system. All these algorithms are recent and represent the state-of-the-art in DeepFake detection (two of the algorithms are from the top-performers of the Global DeepFake Detection Challenge). DeepFake-o-meter is designed to be an open-architecture, which can be augmented by incorporating more detection methods over time. We describe the API structures that are needed for third-party developers of DeepFake detection algorithms to have their method integrated into DeepFake-o-meter.

\section{Platform Design}
\begin{figure*}[t]
    \centering
\includegraphics[width=0.95\linewidth]{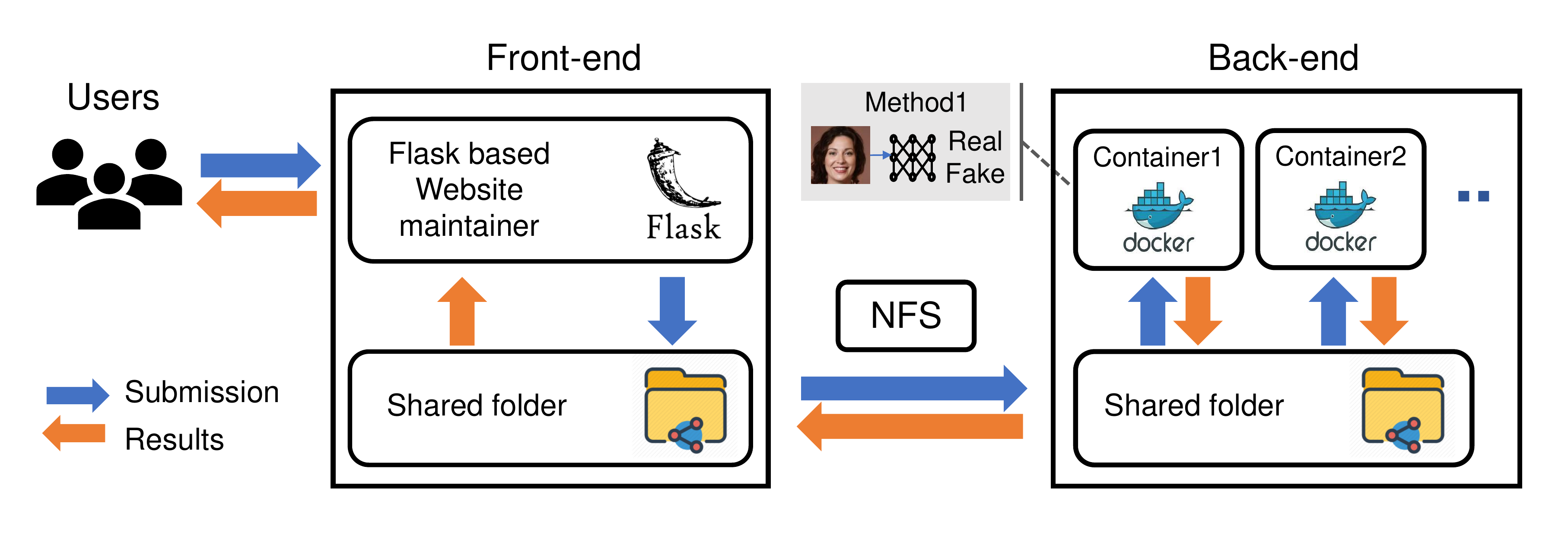}
\caption{The overview of the platform architecture.}
\label{fig:stru}
\end{figure*}

This section describes the architecture design of deepfake-o-meter platform. Our platform is composed by three components: {\em Front-end}, {\em Back-end} and {\em Data synchronizing}. The front-end is the website portal to interact with users. The back-end is the core component of this platform, which calls corresponding detection methods to analyze the submitted videos and the data synchronizing is the protocol for exchanging the interested data between front-end and back-end. The overview of the platform architecture is illustrated in Fig.\ref{fig:stru}. 

\begin{figure}[t]
    \centering
    \includegraphics[width=\linewidth]{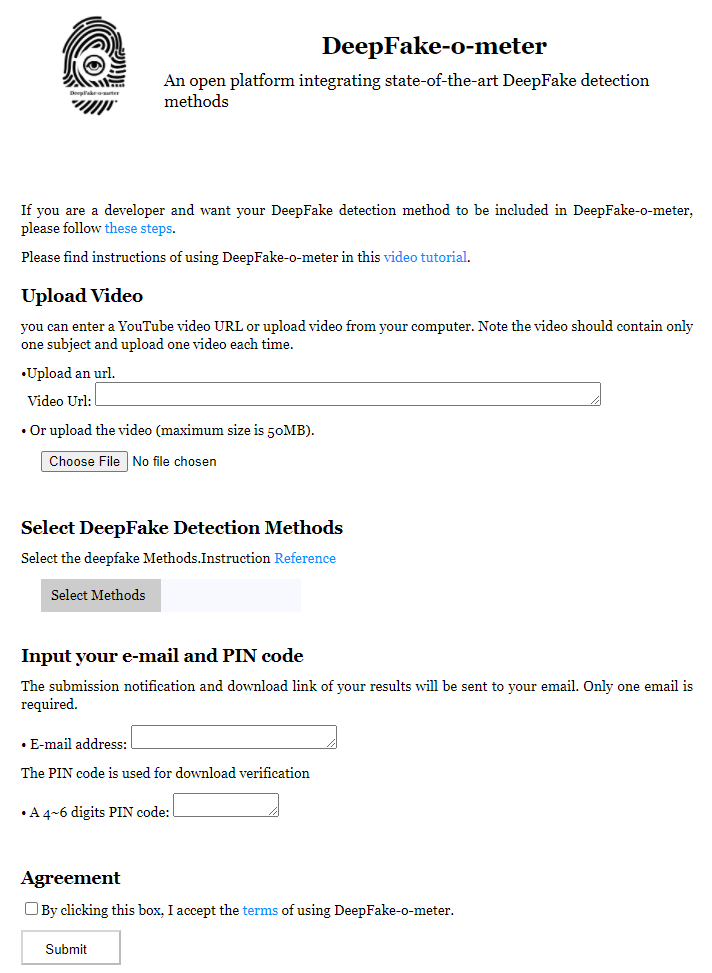}
    \caption{The illustration of the front-end interface.}
    \label{fig:front-end}
\end{figure}

\subsection{Front-end}
In order to interact with users, we develop a website to instruct users to submit their interested videos. Fig.\ref{fig:front-end} shows the illustration of the front-end interface. The steps for users to submit videos are as following:

\begin{enumerate}
    \item Uploading a video either from local machine or using a video URL. Note the maximum video size is constraint to $50$ MB in order to maintain a stable and quicker response;
    \item Selecting the desired deepfake detection methods;
    \item Inputting user's email address and a 4-6 digits PIN code. Note all the subsequent responses including notification and analyzed results will be sent into the provided email address. The PIN code is used for verification for the analyzed results downloading;
    \item The submitted video with other information will sent to the back-end after clicking the submit button.
\end{enumerate}

To construct the front-end, we utilize a python based package {\em Flask}\footnote{\url{https://flask.palletsprojects.com/en/1.1.x/}} as the website maintainer. Flask is a lightweight Web Server Gateway Interface (WSGI) framework, depending on the Jinja template engine and the Werkzeug WSGI toolkit. It is a widely used third-party python library for developing web applications. Flask can take over the routing between different web pages and also the service logic after submission, such as email or PIN code validation and packaging the submission under certain requirements.  


\subsection{Back-end}
The back-end is a computation server mainly for performing deepfake detection methods. In this section, we will describe the key design, returned results and integrated deepfake detection methods respectively.

\subsubsection{Key design}
Once the user submits video from the front-end, the back-end starts to call corresponding detection methods for the submitted video. However, different detection methods depend on different environment settings and different detection methods are designed using different programming styles. Therefore, we design an unified framework to integrate the mainstream deepfake detection methods. Specifically, our framework has two major designs, {\em Container} and {\em Coding structure} to handle the diversity of environment and programming styles respectively.


\smallskip
\noindent{\bf Container.}
We know virtual machines are the first generation tools to solve the environmental conflict problem on a single machine. However, due to the heavy resources occupation, redundant operation and slow startup, virtual machines are replaced by Containers, which can isolate the process without creating a simulate operating system. 
Docker\footnote{\url{https://www.docker.com/}}  is the most popular container solution currently, allowing developers to package their applications and dependent environment into a portable container, which can be run on any other machines. 
To freely run each method, we independently create docker image for each detection method. 

\smallskip
\noindent{\bf Coding structure}
In order to maintain each method efficiently, we design a coding structure that can provide an interface for each method to follow. Specifically, we design a base class containing four basic functions, named \textit{run},  \textit{crop\_face}, \textit{preproc}, \textit{postproc}, \textit{get\_softlabel} and \textit{get\_hardlabel}. 

\begin{enumerate}
    \item[-] \textit{run}: This is the entrance function to process an input image. The input argument is an image and output is the detection score. Given the input image, this function will internally call the functions \textit{crop\_face}, \textit{preproc}, \textit{postproc}, \textit{get\_softlabel}, \textit{get\_hardlabel} in sequel.  

    \item[-] \textit{crop\_face}: Since many methods require to extract the face area from the input image before prediction, this function provides an interface to wrap up the face extraction process. This function is optional.
    
    \item[-] \textit{preproc}: After face extraction, many methods apply pre-processing operations to the input face, such as changing the channel order or color space. Therefore, the pre-processing operations can be put in here. This function is also optional.
    
    \item[-] \textit{get\_softlabel}: This function takes as input the prepossessed face and outputs the confidence score (soft label). Less score denotes the face is faker. The details of calling specific detection methods are wrapped here.
    
    \item[-] \textit{get\_hardlabel}: Based on the soft label, this function assigns the input to real or fake label.
\end{enumerate}

The code structure will also be exposed to the researchers who would like to integrate new methods into platform. The researchers can follow the structure to split their codes into different functions accordingly.

\subsubsection{Returned results}
\begin{figure*}[th!]
    \centering
\includegraphics[width=0.9\linewidth]{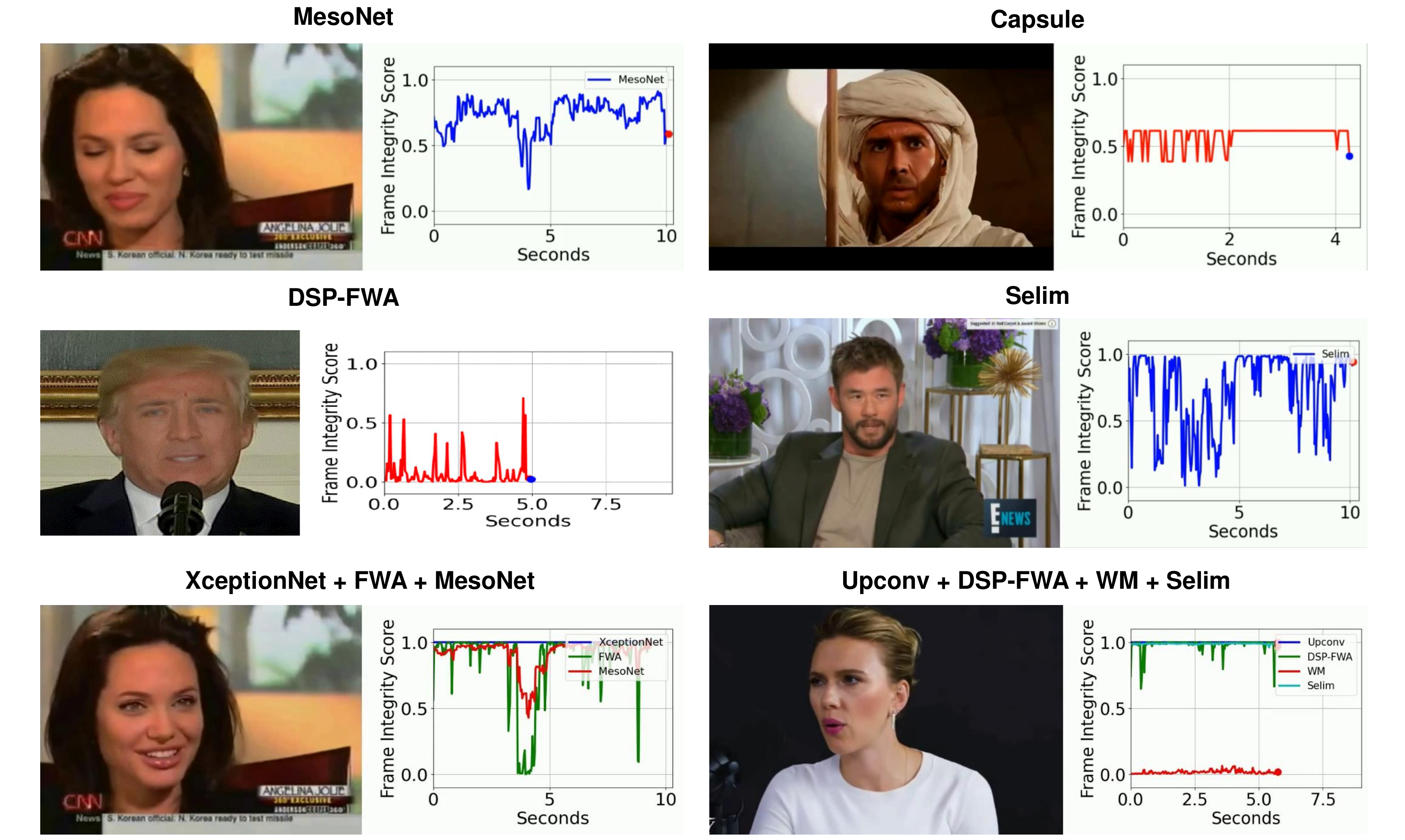}
\caption{Examples of generated results.}
\label{fig:results}
\end{figure*}

The formatting of returned results is also an important point. For better visualization to users, we curve the score of each face along with the corresponding frame and save the prediction of each frame to a video. Besides the visualization, we also sort the score along all frames and calculate the Area Under Curve (AUC) score. The results will be zipped together and sent back to the front-end for user to download. Fig.\ref{fig:results} illustrates several examples of the returned results. The left part is submitted video and right part plots the corresponding score. Note our platform supports to run several methods at the same time, thus the bottom two examples contains multiple curves.

\subsubsection{DeepFake detection methods} Our platform integrates the following deepfake detection methods into this platform. 

\begin{enumerate}
    
    \item {\bf MesoNet} \cite{afchar2018mesonet} is a self-designed CNN model that focuses on the mesoscopic properties of images. They provide two variants of MesoNet, namely,  {\em Meso4} and {\em MesoInception4}. {Meso4 uses conventional convolutional layers, while MesoInception4 is based on the more sophisticated Inception modules \cite{szegedy2015going}.} We integrate {\em MesoInception4} into the platform.  
    
    \item {\bf FWA} \cite{li2019exposing} is based on  ResNet-50 \cite{he2016deep} which detects DeepFake videos by exposing the face warping artifacts due to the resizing and interpolation operations.

    \item {\bf VA} \cite{matern2019exploiting} targets the visual artifacts in the face organs such as eyes, teeth and facial contours of the synthesized faces. Two variants of this method are provided: VA-MLP and VA-LogReg. VA-MLP is based on a crafted CNN, and VA-LogReg uses a simpler logistic regression model. We integrate VA-MLP into the platform
    
    \item {\bf Xception} \cite{roessler2019faceforensics++} comes with FaceForensics++ dataset. It corresponds to a DeepFake detection method based on the XceptionNet model \cite{chollet2017xception}. This method provides three variants: {{\em Xception-raw}, {\em Xception-c23} and {\em Xception-c40}. {\em Xception-raw} are trained on raw videos, while {\em Xception-c23} and {\em Xception-c40} are trained on compressed videos with different degrees, respectively.} We integrate {\em Xception-c23} into the platform.
    
     \item {\bf ClassNSeg} \cite{nguyen2019multi} is another CNN based DeepFake detection method that is formulated to a multi-task learning problem to imultaneously detect forgery images and segment manipulated areas.
    
    \item {\bf Capsule} \cite{nguyen2019capsulev2} employs the VGG19 \cite{simonyan2014very} capsule structure \cite{sabour2017dynamic} as the backbone architecture for DeepFake classification.
    
    \item {\bf DSP-FWA} is a further improved method based on FWA, which incorporates a spatial pyramid pooling (SPP) module \cite{he2015spatial} to better tackle the variations of face resolutions.
    
     \item {\bf CNNDetection} \cite{wang2020cnn} utilizes a standard image classifier trained on only ProGAN \cite{karras2018progressive} , finding it  generalizes surprisingly well to unseen architectures, datasets, and training methods.
     
    \item {\bf Upconv} \cite{durall2020upconv} argues that common up-sampling methods (upconvolution or transposed convolution) lack the ability to reproduce spectral distributions of natural training data correctly. They take $2D$  amplitude spectrum as feature and utilize  a basic SVM classifier. 

    \item {\bf WM} ensembles two WS-DAN \cite{hu2019see} models (with EfficientNet-b3 \cite{tan2019efficientnet} and Xception \cite{chollet2017xception} feature extractors, respectively) and a Xception classifier to produce per-face predictions.  

    \item {\bf Selim} utilizes state-of-the-art encoder, EfficientNet B7, pretrained with ImageNet \cite{imagenet_cvpr09} and noisy student \cite{xie2020self}, and uses a heuristic way to select 32 frames for each video to average predictions.

\end{enumerate}

The summary of each detection method with code repositories is given in Table \ref{table:methods_stat}.

\begin{table*}[t]
\small
\centering
\caption{\small Summary of integrated DeepFake detection methods. See texts for more details.}
  \begin{tabular}{|c|c|c|c|c|}
    \hline
    Methods & Repositories & Release Date \\
    \hline
    \hline
    MesoNet \cite{afchar2018mesonet} & \url{https://github.com/DariusAf/MesoNet} & 2018.09 \\
    \hline
    FWA \cite{li2019exposing} & \url{https://github.com/danmohaha/CVPRW2019_Face_Artifacts} & 2018.11 \\
    \hline
    VA \cite{matern2019exploiting} & {\url{https://github.com/FalkoMatern/Exploiting-Visual-Artifacts}} & {2019.01} \\
    \hline
    Xception \cite{roessler2019faceforensics++} & \url{ https://github.com/ondyari/FaceForensics} & 2019.01 \\
    \hline
    ClassNSeg \cite{nguyen2019multi} & \url{https://github.com/nii-yamagishilab/ClassNSeg} & 2019.06 \\
    \hline
    Capsule \cite{nguyen2019capsulev2} & \url{https://github.com/nii-yamagishilab/Capsule-Forensics-v2} & 2019.10 \\
    \hline
    CNNDetection & \url{https://github.com/peterwang512/CNNDetection} & 2019.12 \\
    \hline
    DSP-FWA & \url{https://github.com/danmohaha/DSP-FWA} & 2019.11 \\
    \hline
    Upconv & \url{https://github.com/cc-hpc-itwm/UpConv} & 2020.03 \\
    \hline
    WM & \url{https://github.com/cuihaoleo/kaggle-dfdc} & 2020.07 \\
    \hline
    Selim & \url{https://github.com/selimsef/dfdc_deepfake_challenge} & 2020.07 \\
    \hline
    
  \end{tabular}
  \label{table:methods_stat}
\end{table*}

\subsection{Data Synchronizing}
This section describes the scheme of data synchronizing between front-end and back-end. To enable data sharing between two machines, we utilize the Network File System (NFS) technology. NFS is a distributed file system protocol that can mount remote directories from client to the server. NFS provides a simple and quick way to visit remote systems through the network. For our platform, we need to set up two shared folders. The first one aims to synchronize the data, \ie, user submitted videos and other information such as email address, from font-end to the back-end. The second one is used to share the detection results of user's submitted videos from the back-end to the front-end, see Fig.\ref{fig:stru}. 

\section{Conclusion}
In this work, we describe an open platform, known as DeepFake-o-meter, for DeepFake detection. This platform is composed by front-end and back-end. The front-end is a web application to interact with users and back-end perform corresponding detection methods on submitted videos. The platform integrates more than 10 state-of-the-art detection methods and it also provides interfaces for researchers to incorporate their method into the platform.

For future works, we will continue integrate more DeepFake detection methods into the platform. Furthermore, we will study the use of multi-GPU platform to accelerate the analysis process. We will also augment the APIs so as to accommodate more general detection methods for other media formats (still images and audio signals).  

\smallskip
\noindent{\bf Acknowledgment}. This work is partly supported by the research project of National Science Foundation (no. IIS-2008532). 

\bibliographystyle{IEEEtran}
\bibliography{ref}

\end{document}